\documentclass[10pt,twocolumn,letterpaper]{article}

\usepackage[pagenumbers]{cvpr} 

\usepackage[dvipsnames]{xcolor}

\usepackage{bbm}
\usepackage{amssymb,mathtools}
\usepackage{multirow}
\usepackage{colortbl}
\usepackage{soul}
\usepackage{comment}
\usepackage{makecell}
\usepackage{pifont}
\usepackage{wrapfig}
\usepackage{subcaption}

\newcommand{\cmark}{\ding{51}}%
\newcommand{\xmark}{\ding{55}}%

\definecolor{cvprblue}{rgb}{0.21,0.49,0.74}
\usepackage[pagebackref,breaklinks,colorlinks,citecolor=cvprblue]{hyperref}
\usepackage[accsupp]{axessibility}

\newcommand{\authorskip}{\hspace{7.5mm}}
\newcommand{\customfootnote}[1]{%
  \begingroup
  \renewcommand{\thefootnote}{}
  \footnote{#1}%
  \endgroup
}
\definecolor{baselinecolor}{gray}{.9}
\newcommand{\baseline}[1]{\cellcolor{baselinecolor}{#1}}

\newcommand{\modelname}[0]{TE-TAD}

\title{\vspace{-2pt}\modelname{}: Towards Full End-to-End Temporal Action Detection via Time-Aligned Coordinate Expression\vspace{-7pt}}

\author{
Ho-Joong Kim$^1$ \authorskip Jung-Ho Hong$^1$ \authorskip  Heejo Kong$^2$ \authorskip
 Seong-Whan Lee$^1$$^*$ \\[2.5pt]
 \normalsize $^1$Dept. of Artificial Intelligence, Korea University, Seoul, Korea\\[0.75pt]
 \normalsize $^2$Dept. of Brain and Cognitive Engineering, Korea University, Seoul, Korea\\[-1pt]
 {\tt\small \{hojoong\_kim, jungho-hong, hj\_kong, sw.lee\}@korea.ac.kr} \vspace{-8pt}\\
}

\begin{document}
\maketitle
\begin{abstract}
\vspace{-5pt}
In this paper, we investigate that the normalized coordinate expression is a key factor as reliance on hand-crafted components in query-based detectors for temporal action detection (TAD).
Despite significant advancements towards an end-to-end framework in object detection, query-based detectors have been limited in achieving full end-to-end modeling in TAD.
To address this issue, we propose \modelname{}, a full end-to-end temporal action detection transformer that integrates time-aligned coordinate expression.
We reformulate coordinate expression utilizing actual timeline values, ensuring length-invariant representations from the extremely diverse video duration environment.
Furthermore, our proposed adaptive query selection dynamically adjusts the number of queries based on video length, providing a suitable solution for varying video durations compared to a fixed query set.
Our approach not only simplifies the TAD process by eliminating the need for hand-crafted components but also significantly improves the performance of query-based detectors.
Our \modelname{} outperforms the previous query-based detectors and achieves competitive performance compared to state-of-the-art methods on popular benchmark datasets. Code is available at: \href{https://github.com/Dotori-HJ/TE-TAD}{https://github.com/Dotori-HJ/TE-TAD}
\end{abstract}

\customfootnote{*Corresponding author}
\vspace{-0.5cm}
\section{Introduction}
\label{sec:introduction}

Temporal action detection (TAD) plays an essential role in video understanding and its numerous real-world applications, such as video surveillance, video summarization, and video retrieval.
TAD aims to recognize and localize actions within untrimmed video sequences by identifying the class labels with precise start and end times of action instances.
Recently, TAD methods can be mainly divided by three approaches: anchor-based  \cite{BSN,BMN,BC-GNN,GTAN,G-TAD,VSGN,TCANet}, anchor-free  \cite{AFSD,TALLFormer,ActionFormer,TriDet}, and query-based  \cite{RTD-Net,ReAct,TadTR} detector.

\begin{figure}[!t]
\centering
    \includegraphics[width=0.94\linewidth]{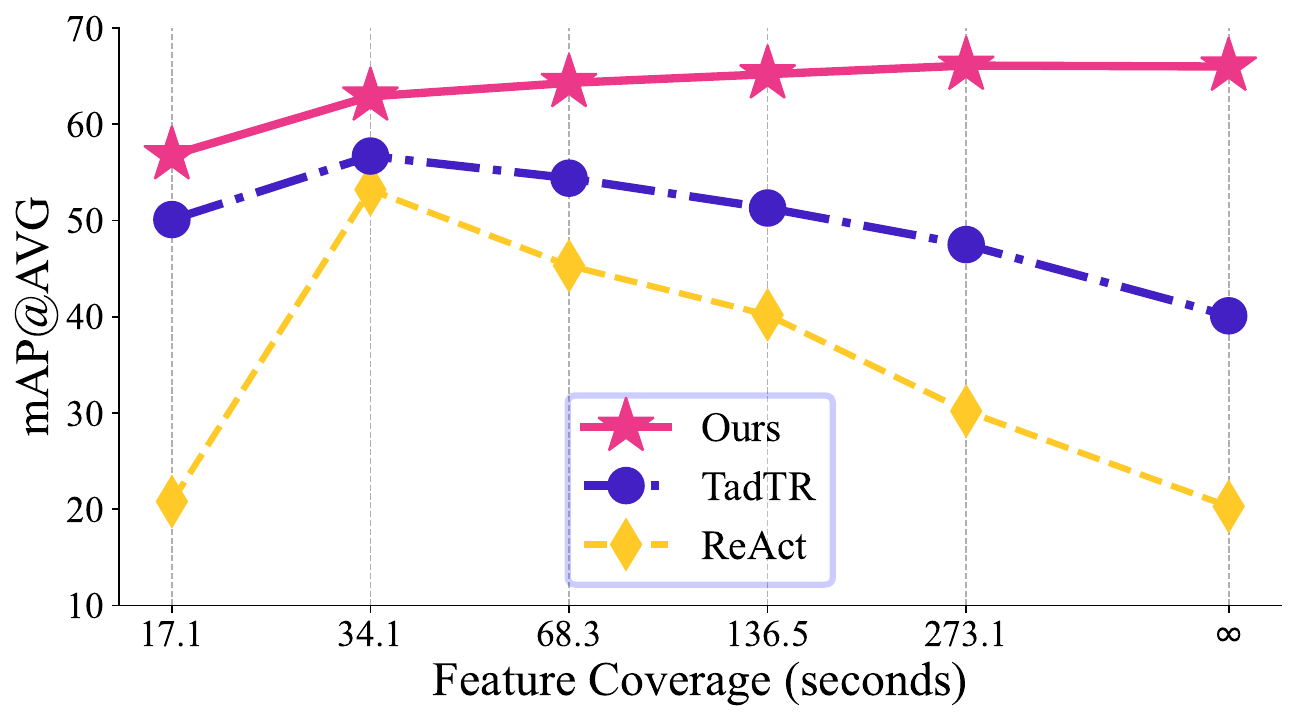}\\ 
    \vspace{-0.25cm}
    \caption{Performance comparison of query-based detectors across various feature coverages on THUMOS14, demonstrating how extending the feature coverage impacts detection performance, as measured by mean Average Precision (mAP@AVG). The full end-to-end setting that without coverage constraints is denoted by $\infty$. }
    \vspace{-0.6cm}
    \label{fig:motivation}
\end{figure}

Query-based detectors, inspired by DETR \cite{DETR}, have attracted interest because of their potential to eliminate reliance on hand-crafted components, such as the sliding window and non-maximum suppression (NMS).
This potential derives from adopting a set-prediction mechanism, which aims to provide an end-to-end detection process by utilizing a one-to-one matching paradigm.
Despite these advantages, query-based detectors encounter significant challenges in two aspects: (1) they show decreased performance when dealing with extended temporal coverage, often making them a less favorable option compared to anchor-free detectors, and (2) due to limited extended temporal coverage, the reliance on the sliding window approach leads to redundant proposals and necessitates the use of NMS.

To demonstrate these issues, we conduct experiments by increasing the feature coverages on existing query-based detectors.
Fig.~\ref{fig:motivation} demonstrates the change in performance across various feature coverage.
Except for unrestricted feature coverage denoted by $\infty$, feature coverage is calculated from the window size of the sliding window method.
As shown in the graph, even though wide feature coverage is beneficial to capturing longer context, the performance of existing query-based detectors diminishes in mean Average Precision (mAP) as feature coverage increments.
This result indicates that there are significant limitations to query-based detectors in scalability to temporal length, despite efforts to address long durations \cite{ActionFormer} in TAD.

\begin{figure}[!t]
\centering
    \begin{subfigure}[c]{0.9\linewidth}
        \centering
        \includegraphics[width=\linewidth]{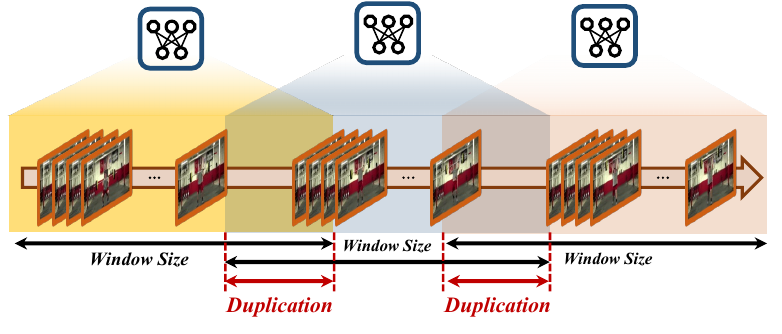}\\ 
        \vspace{-0.15cm}
        \caption{Sliding window}
        \label{fig:window}
        \vspace{0.05cm}
    \end{subfigure}
    \begin{subfigure}[c]{\linewidth}
        \centering
        \includegraphics[width=0.9\linewidth]{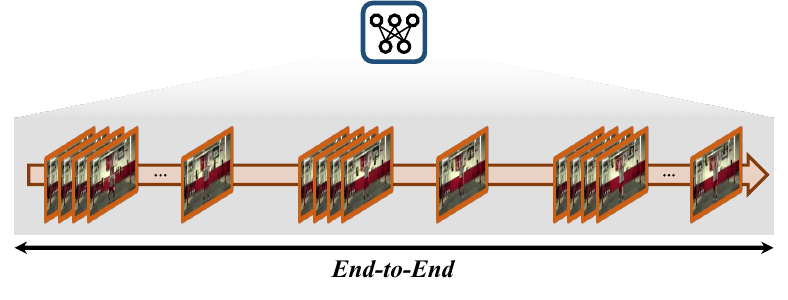} \\
        \vspace{-0.15cm}
        \caption{End-to-end setting}
        \label{fig:e2e}
    \end{subfigure}
    \vspace{-0.3cm}
    \caption{Comparison between sliding window and end-to-end settings. The sliding window generates redundant proposals in the duplicated area.}
    \vspace{-0.4cm}
    \label{fig:window_e2e}
\end{figure}

Furthermore, Fig.~\ref{fig:window_e2e} shows the second issue, illustrating the limitations of the sliding window approach.
As shown in Fig.~\ref{fig:window_e2e}\subref{fig:window}, the sliding window is limited to address long-range duration due to its limited window size.
Moreover, the sliding window approach contains overlapping areas to prevent predictions from being truncated in the middle.
These overlapping areas generate duplicate predictions, necessitating the use of NMS to filter the false positive cases.
This reliance on NMS contradicts the set-prediction goal of minimizing hand-crafted components, thus hindering the achievement of a fully end-to-end TAD.
To address these issues, we investigate why extended temporal coverage adversely affects the performance of query-based detectors (Sec.~\ref{subsec:instability_sensitivity}).
Our investigation reveals that the conventional use of normalized coordinate expressions is a significant factor, disturbing the achievement of a full end-to-end TAD.

In this paper, we propose a full end-to-end temporal action detection transformer that integrates time-aligned coordinate expression (\modelname{}), which reformulates normalized coordinate expression to actual timeline video values.
Our reformulation enables the query-based detector to effectively address length-invariant modeling by avoiding the distortion of the normalizing process, which not only enhances the detection performance but also simplifies the detection process.
Our \modelname{} stabilizes the training process of the query-based detector when dealing with extended videos; our approach shows significant improvements and completely removes the reliance on hand-crafted components such as the sliding window and NMS.
Furthermore, we introduce an adaptive query selection that effectively addresses various video lengths, dynamically adjusting the number of queries in response to the temporal length of each video.
In contrast to relying on a fixed set of queries, our \modelname{} provides a suitable approach to process diverse video lengths.
Our approach shows significant improvements compared to previous query-based detectors and achieves competitive performance with state-of-the-art methods on popular benchmark datasets: THUMOS14 \cite{THUMOS14}, ActivityNet~v1.3 \cite{ActivityNet}, and EpicKitchens \cite{EpicKitchens}.

\noindent Our contributions are summarized as three-fold:
\begin{itemize}
    \item We propose a full end-to-end temporal action detection transformer that integrates time-aligned coordinate expression (\modelname{}), which preserves the set-prediction mechanism and enables a full end-to-end modeling for TAD by eliminating the hand-crafted components.
    \item Our approach introduces a length-invariant mechanism to query-based detectors, significantly improving scalability in handling varying lengths of videos.
    \item Our \modelname{} significantly outperforms the previous query-based detectors and achieves competitive performance compared to state-of-the-art methods, even without hand-crafted components such as the sliding window and NMS.
\end{itemize}

\section{Related Work}
\label{sec:related_work}

\noindent \textbf{Action Recognition}
Action Recognition is a foundational task in video understanding, categorizing video sequences into distinct action classes.
Notable models include I3D \cite{I3D}, which enhances the inception network with 3D convolutions, and R(2+1)D \cite{R2+1D}, separating 3D convolutions into 2D spatial and 1D temporal parts for efficient processing. TSP \cite{TSP} introduces temporal channel shifting for effective temporal modeling without extra computation. VideoSwin \cite{VideoSwin}, employing the SwinTransformer \cite{Swin} architecture, excels in complex video data recognition.
These models serve as backbones for extracting video features, directly impacting performance in subtasks like TAD.

\noindent \textbf{Anchor-based Detector}
Anchor-based detectors \cite{BSN,BMN,BC-GNN,GTAN,G-TAD,VSGN,TCANet} leverage predefined anchor boxes to generate action proposals.
These hand-designed anchors hinder the diverse range of action instances because of a lack of flexibility in the localization of the action instances.
This approach inherently limits and necessitates additional post-processing steps to discard redundant proposals because they model the one-to-many assignment training strategy.

\noindent \textbf{Anchor-free Detector}
Anchor-free detectors \cite{AFSD,TALLFormer,ActionFormer,ASL,MENet,TriDet} offer more flexibility in action instance localization compared to anchor-based detectors by adopting an asymmetric modeling approach.
For instance, ActionFormer \cite{ActionFormer} significantly enhances TAD performance by employing a transformer-based architecture that captures long-range video dependencies.
TriDet \cite{TriDet} demonstrates superior performance in TAD by employing a trident prediction scheme and their proposed convolution-based architecture.
Despite their improvements, anchor-free detectors rely on hand-crafted components to remove redundant proposals using NMS because they adopt a one-to-many assignment training manner.
In contrast, our approach directly addresses a one-to-one matching scheme, eliminating the use of NMS.

\noindent \textbf{Query-based Detector}
Query-based detectors, inspired by DETR \cite{DETR}, introduce a set-prediction mechanism, thereby reducing the reliance on hand-crafted components, which ideally prevents the need for NMS.
However, existing query-based detectors still require NMS because their model design inherently breaks the one-to-one matching paradigm.
RTD-Net \cite{RTD-Net} utilizes a one-to-many matching to mitigate the slow convergence issue associated with the detection transformer.
This approach inherently breaks the one-to-one assignment.
ReAct \cite{ReAct} modify the decoder's self-attention, called relational attention, only adopt self-attention between their defined relations.
This partial adoption of the decoder's self-attention disturbs the set-prediction mechanism because they cannot capture the whole context of queries.
Furthermore, previous query-based detectors \cite{RTD-Net,ReAct,TadTR} adopt the sliding window method that contains overlapping areas, causing redundant proposals.
Furthermore, TadTR \cite{TadTR} deals with one-to-many matching at training loss. TadTR employs cross-window fusion (CWF), which applies NMS to overlapping areas to remove redundant proposals.
In contrast, our approach entirely preserves the one-to-one matching paradigm, which enables a full end-to-end modeling.
\section{Our Approach}
\subsection{Overview}
\begin{figure*}[!t]
\centering
    \begin{tabular}{c}
        \includegraphics[width=\linewidth]{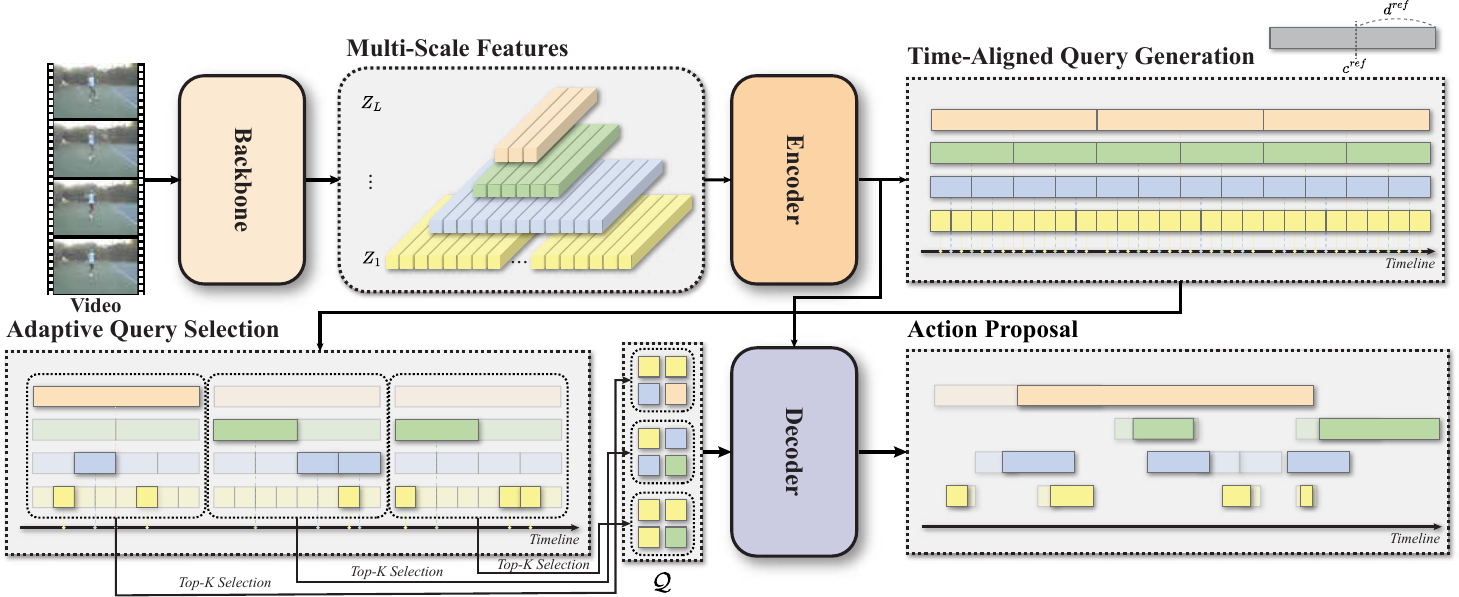}\\
    \end{tabular}
    \vspace{-0.25cm}
    \caption{Overview of the \modelname{}. Starting with video input, the architecture processes through a backbone for feature extraction, generating multi-scale features $Z$. These are encoded and subsequently passed through an adaptive query selection, aligning with the video timeline for initial query generation. The decoder refines these queries layer-by-layer, culminating in the refinement of action proposals.}
    \vspace{-0.4cm}
    \label{fig:model}
\end{figure*}

In this section, we first discuss about the limitations of existing query-based detectors in TAD, focusing on the normalized coordinate expression.
The normalized coordinate expression, used in existing models, causes matching instability and sensitivity, especially in extended video scenarios.
Subsequently, to introduce our \modelname{}, we describe the reformulation of normalized coordinate expression to timeline coordinate expression and adaptive query proposals to ensure the length-invariant modeling.
The overall architecture of \modelname{} is illustrated in Fig.~\ref{fig:model}.

\subsection{Preliminary}
\label{subsec:preliminary}
Let $X \in \mathbb{R}^{T_0 \times C}$ denote the video feature sequence extracted by the backbone network, where $T_0$ is the temporal length of the features, and $C$ is the dimension of the video feature.
Each element in the video feature sequence represented as $X = \{ x_t \}_{t=1}^{T_0}$, corresponds to a snippet at timestep $t$, with each snippet comprising a few consecutive frames.
These snippets are processed using a pre-trained backbone network such as I3D \cite{I3D} or SlowFast \cite{SlowFast}.
Each video contains numerous action instances, and each action instance contains start and end timestamps $s$ and $e$, along with its action class $c$.
Formally, the set of action instances in a video is represented as $\mathcal{A} = \{(s_n, e_n, c_n)\}_{n=1}^{N}$, where $N$ is the number of action instances, and $s_n$ and $e_n$ are the start and end timestamps of an action instance, respectively, and $c_n$ is its action class.
The main goal of TAD is to accurately predict the set of action instances $\mathcal{A}$ for any given video.
Previous query-based detectors \cite{ReAct,RTD-Net,TadTR} typically compute predicted values of center $\hat{c}$ and width $\hat{d}$ using a sigmoid function ($\sigma$).
Consequently, existing models decode predicted start $\hat{s}$ and end $\hat{e}$ timestamps by $\sigma(\hat{c}) - \sigma(\hat{d})$ and $\sigma(\hat{c}) + \sigma(\hat{d})$ to start and end timestamps, respectively.

\begin{figure}[!t]
    \centering
    \begin{subfigure}[c]{0.8\linewidth}
        \centering
        \includegraphics[width=\linewidth]{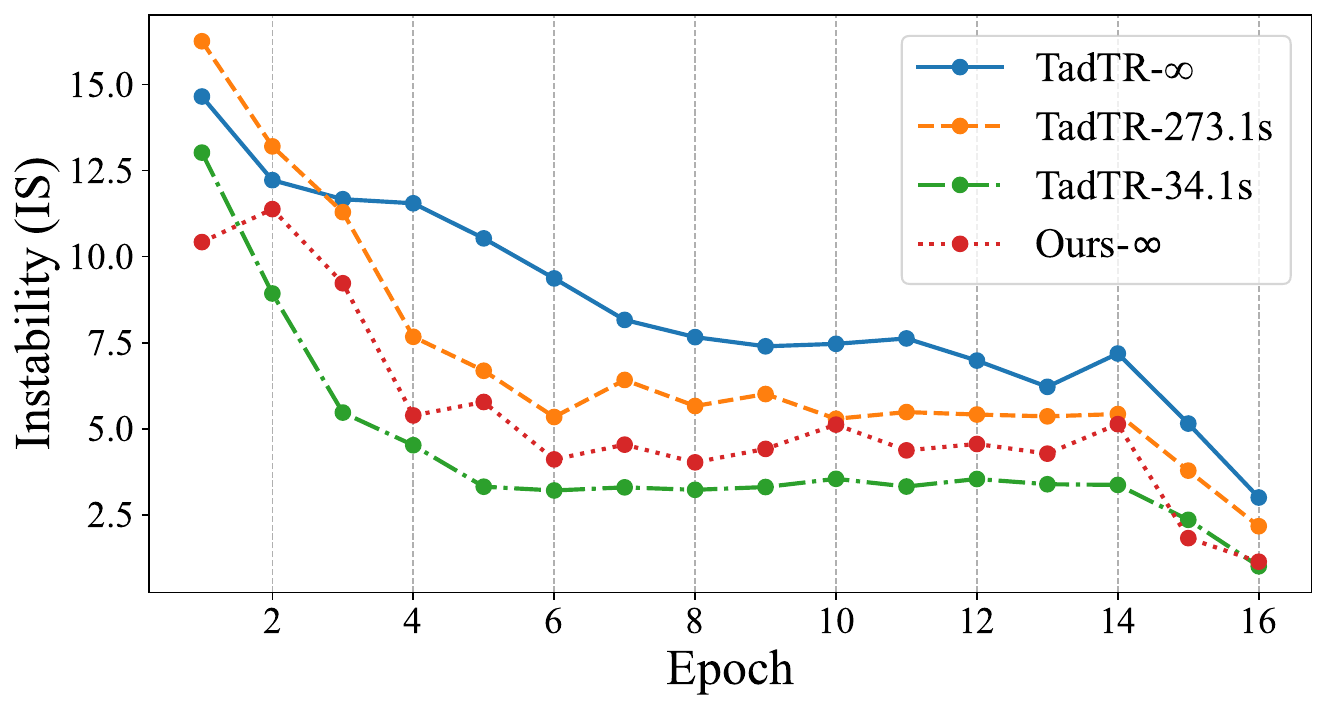}
        \vspace{-0.5cm}
        \caption{}
        \label{fig:is}
    \end{subfigure}
    \begin{subfigure}[c]{0.8\linewidth}
        \centering
        \includegraphics[width=\linewidth]{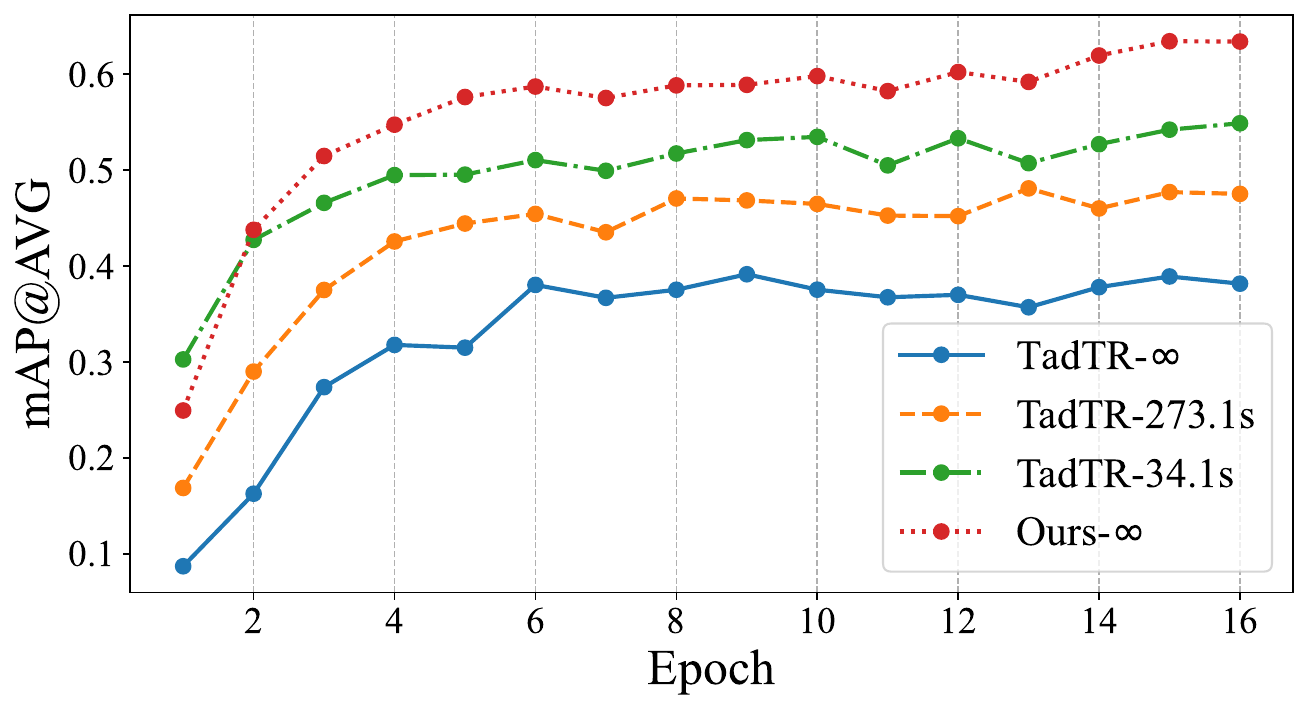}
        \vspace{-0.5cm}
        \caption{}
        \label{fig:convergence}
    \end{subfigure}
    \vspace{-0.3cm}
    \caption{Comparative analysis of instability and detection performance on THUMOS14: (a) variance in instance matching quantified by IS; (b) the performance change in mAP. Feature coverage lengths are denoted by the values following the dash (-). The symbol $\infty$ denotes an unrestricted end-to-end setting.}
    \label{fig:is_convergence}
    \vspace{-0.4cm}
\end{figure}

\subsection{Exploring Key Issues}
\label{subsec:instability_sensitivity}
\noindent \textbf{Matching Instability}
As discussed in \cref{sec:introduction}, extended video lengths significantly influence the performance of existing query-based methods.
To investigate this issue, we conduct a comparative study on the instability (IS) \cite{DN-DETR} and the detection performance across different feature coverage scenarios on THUMOS14.
Here, IS is a quantitative measurement of the inconsistency of matching during the training process.
For query-based detectors, fluctuations in matched targets compel the model to be learned from different values for the same input, leading to performance degradation.
In the TadTR setting, smaller window sizes have more steps per epoch due to generating more sliding windows, whereas larger window sizes yield fewer steps per epoch.
This mismatch in the number of steps results that with fewer updates per epoch, there is inherently less change to the model, which shows lower instability.
For a fair comparison, we match the number of iteration steps per epoch, aligning with TadTR-34.1s (original TadTR).

Fig.~\ref{fig:is_convergence} illustrates the instability and detection performance across diverse feature coverage.
As feature coverage increases, we observe a rise in IS across the training epoch, indicating less stable instance matching, which leads to a decline in detection performance.
This analysis underscores the challenge of maintaining consistent learning when with extended temporal lengths.
Furthermore, a direct comparison between TadTR models and our method reveals that Ours-$\infty$ maintains a level of stability comparable to TadTR-34.1s.
This indicates that our approach significantly stabilizes the training process relative to TadTR-$\infty$.
Moreover, our method demonstrates similar levels of matching instability yet shows significant improvements under more challenging conditions for matching problems, even when compared to models with shorter feature coverage like TadTR-34.1s and TadTR-273.1s.

\begin{figure}[!h]
    \centering
    \vspace{-0.2cm}
    \begin{subfigure}[c]{0.492\linewidth}
        \centering
        \includegraphics[width=\linewidth]{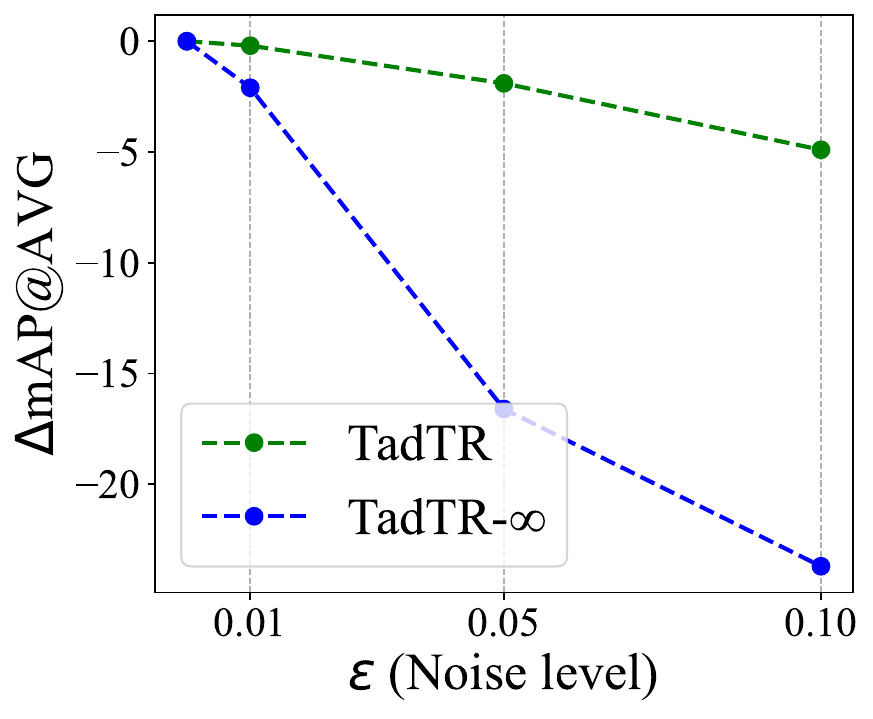}
        \caption{$\sigma(\hat{c} + \epsilon)$}
        \label{fig:center_center}
    \end{subfigure}
    \begin{subfigure}[c]{0.492\linewidth}
        \centering
        \includegraphics[width=\linewidth]{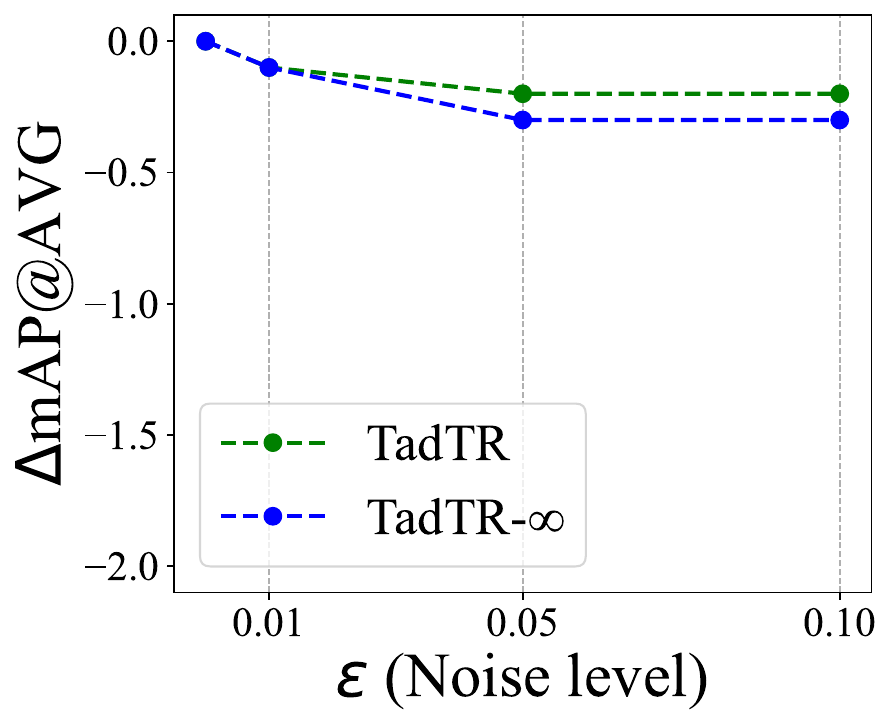}
        \caption{$\sigma(\hat{d} + \epsilon)$}
    \end{subfigure}
    \vspace{-0.25cm}
    \caption{Analysis of noise tolerance on predicted value. The noise level $\epsilon$ sampled from uniform distribution and injected before the \textit{sigmoid} function.}
    \vspace{-0.3cm}
    \label{fig:center}
\end{figure}

\noindent \textbf{Sensitivity to Localize Action Instances}
To further explore this issue, we investigate the sensitivity of model prediction measured by adopting minor perturbations in predictions of localization.
We inject a small scale of noise sampled from a uniform distribution before adopting the \textit{sigmoid} function that normalizes coordinates within a [0, 1] range.
Fig.~\ref{fig:center} shows that even minimal noise injection significantly affects the prediction when applying the center prediction value.
The noise $\epsilon \sim \textrm{Uniform}(-\alpha, \alpha)$ is sampled from the uniform distribution.
As illustrated in Fig.~\ref{fig:center}\subref{fig:center_center}, the model’s sensitivity to small shifts is evident from the significant decline in $\Delta$mAP@AVG upon introducing noise to the center predicted value.
Notably, with noise levels of $\pm0.01$ and $\pm0.1$, after processing through the \textit{sigmoid} function ($\sigma$), the maximal shifts are confined within $\pm0.0025$ and $\pm0.025$ in the normalized coordinate space, respectively.
These findings highlight that even minor output variations amplify in extended videos, leading to considerable drops in performance due to the heightened sensitivity in the normalized coordinate framework.
To address this issue, we introduce a time-aligned coordinate expression that is not normalized, thereby ensuring independence from video length and reducing sensitivity.

\subsection{\modelname{}}
This part describes our \modelname{} for a full end-to-end TAD.
We adopt the TadTR \cite{TadTR} architecture as a baseline method, including encoder, decoder, and temporal deformable attention architecture.
Starting with the baseline, we mainly address three aspects:
(1) adopting multi-scale and two-stage methods from the previous methods in object detection to bridge the performance gap between query-based and anchor-free detectors,
(2) reformulating coordinate expression utilizing the actual time values to address extended video length in an end-to-end setting, and
(3) proposing an adaptive query selection that dynamically adjusts the number of queries based on the diverse length of videos.

\noindent \textbf{Embedding \& Multi-Scale Features}
We project input features $X$ using a single convolutional neural network to align them with the dimension of the transformer architecture.
The projection maps the input feature $X$ to the embedded feature $Z_1 \in \mathbb{R}^{D \times T_1}$, where $D$ denotes the channel dimension of the encoder and decoder transformer architecture.
The temporal length $T_1$ of the embedded features $Z_1$ remains the same as the original $T_0$.
Subsequently, following previous approaches \cite{ActionFormer,TriDet}, we incorporate multi-scale generation to effectively address varying lengths of actions.
Unlike utilizing a transformer \cite{ActionFormer}, we employ a single convolution layer with a stride of 2 to produce features at each scale level as follows:
\begin{equation}
    \small Z_{l} = \textrm{LayerNorm}_l(\textrm{Conv}_l(Z_{l-1}))\  ,l \in \{2,...,L\},
\end{equation}
where $Z_{l} \in \mathbb{R}^{C \times T_l}$ represents the embedded features at each level $l$, and $L$ denotes the total number of feature levels.
Each subsequent level $l$ has a temporal length $T_l$ that is half of the temporal level of the previous level $T_{l-1}$.
The $\textrm{LayerNorm}_l$ and $\textrm{Conv}_l$ denote $l$-th layer normalization and convolutional neural networks, respectively.
We do not apply any activation function in this process to deliver the raw feature representations to the transformer detector.

\noindent \textbf{Time-Aligned Query Generation}
Our method follows a two-stage approach \cite{Deformable-DETR,DINO} that generates initial action proposals using the transformer encoder.
The previous two-stage approach \cite{Deformable-DETR} provides the reference to the encoder's outputs.
The transformer encoder predicts a binary foreground score, $p^{(0)}$, and segment offsets $\Delta c^{(0)}$ and $\Delta d^{(0)}$ to refine segments based on reference, for each time $t$ and level $l$.
We define reference for center $c^{\text{ref}}$ and width $d^{\text{ref}}$ predictions, aligning with the real timeline of the video.
For each scale level $l$, the reference for the center is computed as follows:
\begin{equation}
\small c^{\text{ref}} = \left \{ t \times \frac{f}{w \times 2^{l-1}} + \frac{w \times 2^{l-1}}{2} \right \}_{t=1}^{T_l},~l \in \{1,2,\dots, L\},
\end{equation}
where $f$ denotes the frame-per-second rate of the video, and $w$ represents step size for feature extracting that indicates how many frames to step when extracting features.
The factor $2^{l-1}$ fits the temporal lengths to the embedded features from the multi-scale generation, corresponding with the actual timeline.
Subsequently, the reference for the width is computed as follows:
\begin{equation}
    \small d^{\text{ref}} = \alpha \cdot f \times 2^{l-1},~l \in \{1,2,\dots, L\},
\end{equation}
where $\alpha$ is the base scale for encoder proposals, which adjusts the length of the reference width length.
Consequently, the time-aligned queries are decoded as follows:
\begin{equation}
\small (\hat{c}^{(0)}, \hat{d}^{(0)}) = (c^{\text{ref}} + \Delta c^{(0)} \cdot d^{\text{ref}}, \exp(\ln(d^{\text{ref}}) + \Delta d^{(0)}))
\end{equation}
where $\hat{c}^{(0)}$ and $\hat{d}^{(0)}$ denote the center and width of the proposals, respectively.
Our approach utilizes the scaling of the center offsets with width $d^{\text{ref}}$ and $\exp$ at the proposals to address the scale-invariant approach.
The decoder then refines these initial locations of queries after the proposed adaptive query selection.

\noindent \textbf{Adaptive Query Selection}
In a conventional two-stage approach in object detection, a fixed top-$k$ selection method based on binary class predictions $p^{(0)}$ is typically employed.
However, this static method may not be optimal for TAD, where the number and duration of action instances within videos vary significantly.
In TAD, there's often a direct correlation between the length of a video and the number of action instances.
Typically, longer videos contain more action instances, while shorter videos have fewer.
This variation presents a challenge for the fixed selection method, which fails to adapt to video content characteristics.

Our method divides videos into sectors of a base length $T_{\text{sector}}$.
We then perform top-$k$ selection for each sector.
This individual selection allows the detector to adapt to the video length, preventing the selection of only some parts when dealing with long videos.
In practice, we redistribute any remaining timesteps to ensure that no sector at the end part of the video is smaller than $T_{\text{sector}}$.
The total number of sectors, $S$, is calculated by dividing the number of the first layer's feature $T_1$ by $T_{\text{sector}}$, applying a floor function.
For each sector $s$, where $s \in \{1, 2, \dots, S\}$, we select the top-$k$ proposals from all levels based on the encoder's binary class scores within that sector.
We define a subset of encoder output scores, $P_s$, for each sector, and then select the top-$k$ proposals from this subset.
The adaptive query selection is represented as follows:
\begin{equation}
\small \mathcal{Q} = \bigcup_{s=1}^{S} \left\{ (\hat{c}_{t,l}^{(0)}, \hat{d}_{t,l}^{(0)}) \mid (t, l) \in \text{indices of top-}K \  \text{in} \  P_s  \right\},
\end{equation}
where $\mathcal{Q}=\{ (\hat{c}_{q}^{(0)}, \hat{d}_{q}^{(0)}) \}_{q=1}^{N_q}$ is the aggregated set of selected queries.
Here, $K$ is the number of queries selected from each sector, and the total number of queries $N_q$ equals the sum of the top-$k$ proposals across all sectors, denoted as $N_q = \sum_{s=1}^{S}{K}$.

\noindent \textbf{Time-Aligned Segment Refinement}
Existing query-based models \cite{ReAct,RTD-Net,TadTR} employ the \textit{sigmoid} function to express normalized coordinates from the range 0 to 1.
Moreover, TadTR \cite{TadTR} utilizes the refining step in the decoder of the transformer using the predicted center and width of each layer.
The layer-wise segment refinement step in the normalized coordinate expression is defined as $\sigma ( \sigma^{-1}(\hat{c}_q^{(n-1)}) + \Delta c_q^{(n)})$ and $\sigma ( \sigma^{-1}(\hat{d}_q^{(n-1)}) + \Delta d_q^{(n)})$ for center prediction and width prediction, respectively.
This refinement step restricts the values within a $[0, 1]$ but enables layer-by-layer updates.
We reorganize the previous segment refinement step without the normalized expression.
In our approach, given selected queries $\mathcal{Q}$ are utilized for segment refinement.
Formally, segment refinement of each layer is as follows:
\begin{equation}
    \small \hat{c}_q^{(n)} = \hat{c}^{(n - 1)}_q + \Delta c^{(n)}_q \cdot \hat{d}^{(n - 1)}_q, \  n \in \{1,2, \dots, L_D\},
\end{equation}
\begin{equation}
    \small \hat{d}_q^{(n)} = \exp \left(\ln(\hat{d}^{(n - 1)}_q) + \Delta \hat{d}^{(n)}_q \right), \ n \in \{1,2, \dots, L_D\},
\end{equation}
where $\hat{c}_q^{(n)}$ and $\hat{d}_q^{(n)}$ represent the predicted outputs of the center point and width from the decoder, respectively, of $n$-th layer for each query.
Similar to the encoder case, we utilize the scaling of the center offsets with width $d^{(n)}$ and exp function at the proposals to address the scale-invariant approach.
The start $\hat{s}^{(L_D)}$ and end $\hat{e}^{(L_D)}$ timestamps for predictions are decoded as $\hat{c}^{(L_D)} - \hat{d}^{(L_D)}$ and $\hat{c}^{(L_D)} + \hat{d}^{(L_D)}$, respectively.
Consequently, the final predicted proposals is defined as $\hat{\mathcal{A}} = (\hat{s}^{(L_D)}, \hat{e}^{(L_D)}, \hat{p}^{(L_D)})$, where $L_D$ is the number of decoder layer, and $\hat{p}^{(L_D)}$ is the prediction of confidence score of last layer of decoder action class.

\subsection{Training and Inference}
\label{sec:training_and_inference}
\noindent \textbf{Training}
We follow the standard bipartite matching loss \cite{DETR}.
The total loss $\mathcal{L}_{total}$ is defined as follows:
\begin{equation}
    \mathcal{L}_{total}(\mathcal{A}, \hat{\mathcal{A}}) = \sum_{i=1}^{N_q}{\mathcal{L}_{match} (\mathcal{A}_i, \hat{\mathcal{A}}_{\pi(i)})},
    \label{eq:total_loss}
\end{equation}
where $\mathcal{L}_{match}$ is the bipartite matching loss that incorporates both the classification probabilities and the distance between the ground truth and the predicted segments, and $\pi$ represent permutation indices obtained by bipartite matching \cite{HungarianMatching}.
This cost function, $\mathcal{L}_{match}$, is a composite of the classification loss, and the regression loss.
For the classification loss, we use the focal loss \cite{FocalLoss}, which effectively addresses the class imbalance issue.
For the regression loss, our model utilizes DIoU \cite{DIoU} and log-ratio distance for the width.
DIoU evaluates the relative center distance and GIoU \cite{GIoU} once, while the log-ratio compares the widths relatively.
Following the previous query-based approaches \cite{TadTR,DETR,Deformable-DETR}, we utilize the auxiliary decoding loss at every decoder layer for a more effective learning process.
Furthermore, we do not apply any other losses without bipartite matching loss, such as actionness loss \cite{TadTR} or action classification enhancement loss \cite{ReAct}.
More detailed descriptions of each loss are described in the supplementary materials.

\noindent \textbf{Inference}
\modelname{} introduces a significant innovation by completely removing the need for common post-processing steps, such as NMS and temporal scaling.
This is a direct consequence of our model's unique ability to work with an end-to-end approach and actual timeline values of the video, streamlining the inference process.
The predictions from the final layer of the decoder $\hat{\mathcal{A}}$ are directly used.
\section{Experiments}
\label{sec:experiments}
\subsection{Setup}
\noindent \textbf{Datasets} We conduct experiments on three datasets: THUMOS14 \cite{THUMOS14}, ActivityNet~v1.3 \cite{ActivityNet}, and EpicKitchens \cite{EpicKitchens}.
THUMOS14 comprises 20 action classes with 200 and 213 untrimmed videos in the validation and test sets, containing 3,007 and 3,358 action instances, respectively.
ActivityNet~v1.3 are large-scale datasets with 200 action classes.
They consist of 10,024 videos for training and 4,926 videos for validation, respectively.
EpicKitchens, a first-person vision dataset, includes two sub-tasks: noun and verb.
It contains 495 and 138 videos with 67,217 and 9,668 action instances for training and test, with 300 and 97 action classes for nouns and verbs, respectively.
These datasets contain diverse actions and scenes, providing a rigorous evaluation setup for our method.

\noindent \textbf{Evaluation Metric}
We follow the standard evaluation protocol for all datasets, utilizing mAP at different intersections over union (IoU) thresholds to evaluate TAD performance.
The IoU thresholds for THUMOS14 and EpicKitchens are set at [0.3:0.7:0.1] and [0.1:0.5:0.1] respectively, while for ActivityNet~v1.3, the results are reported at IoU threshold [0.5, 0.75, 0.95] with the average mAP computed at [0.5:0.95:0.05].

\noindent \textbf{Implementation Details}
To ensure the clarity and focus of our main manuscript, detailed descriptions of the hyperparameters and experimental environments are provided in the supplementary materials.

\begin{table}[!h]
\centering

\resizebox{0.95\linewidth}{!}{
\begin{tabular}{ccccccccc}
\Xhline{2\arrayrulewidth}
\noalign{\smallskip}
\multirow{2.7}{*}{\textbf{Type}} & \multirow{2.7}{*}{\textbf{Method}} & \multirow{2.7}{*}{\textbf{NMS}} & \multicolumn{6}{c}{\textbf{mAP}}\\
\noalign{\smallskip}
\cline{4-9}
\noalign{\smallskip}
& & & 0.3 & 0.4 & 0.5 & 0.6 & 0.7 & Avg.\\
\noalign{\smallskip}
\Xhline{2\arrayrulewidth}
\noalign{\smallskip}

\multirow{6}{*}{\shortstack{Anchor\\-based}}
& BSN \cite{BSN}$^*$ & \cmark & 
53.5 & 45.0 & 36.9 & 28.4 & 20.0 & 36.8\\
& BMN \cite{BMN}$^*$ & \cmark & 
56.0 & 47.4 & 38.8 & 29.7 & 20.5 & 38.5\\
& BC-GNN \cite{BC-GNN}$^*$ & \cmark & 
57.1 & 49.1 & 40.4 & 31.2 & 23.1 & 40.2\\
& G-TAD \cite{G-TAD}$^*$ & \cmark & 
54.5 & 47.6 & 40.3 & 30.8 & 23.4 & 39.3\\
& VSGN \cite{VSGN}$^*$ & \cmark & 
66.7 & 60.4 & 52.4 & 41.0 & 30.4 & 50.2\\
& TCANet \cite{TCANet}$^*$ & \cmark & 
60.6 & 53.2 & 44.6 & 36.8 & 26.7 & 44.3\\

\noalign{\smallskip}
\hline
\noalign{\smallskip}

\multirow{5}{*}{\shortstack{Anchor\\-free}}

& AFSD \cite{AFSD} & \cmark & 
67.3 & 62.4 & 55.5 & 43.7 & 31.1 & 52.0\\
& MENet \cite{MENet}$^\ddagger$ & \cmark & 70.7 & 65.3 & 58.8 & 49.1 & 34.0 & 55.6\\ 
& TALLFormer \cite{TALLFormer}$^\dagger$ & \cmark & 
76.0 & - & 63.2 & - & 34.5 & 59.2\\
& ActionFormer \cite{ActionFormer} & \cmark & 
82.1 & 77.8 & 71.0 & 59.4 & 43.9 & 66.8\\
& TriDet \cite{TriDet} & \cmark & 
\textbf{83.6} & \textbf{80.1} & \textbf{72.9} & \textbf{62.4} & \textbf{47.4} & \textbf{69.3}\\

\noalign{\smallskip}
\hline
\noalign{\smallskip}

\multirow{6}{*}{\shortstack{Query\\-based}}
& RTD-Net \cite{RTD-Net} & \cmark & 
68.3 & 62.3 & 51.9 & 38.8 & 23.7 & 49.0\\
& ReAct \cite{ReAct} & \cmark & 
69.2 & 65.0 & 57.1 & 47.8 & 35.6 & 55.0\\
& Self-DETR \cite{Self-DETR} & $\mathbf{\Delta}$ & 
74.6 & 69.5 & 60.0 & 47.6 & 31.8 & 56.7\\
& TadTR \cite{TadTR} & $\mathbf{\Delta}$ &
74.8 & 69.1 & 60.1 & 46.6 & 32.8 & 56.7\\
& \baseline{\modelname{} (Ours)} & \baseline{\xmark} &
\baseline{81.7} & \baseline{76.6} & \baseline{69.5} & \baseline{59.3} & \baseline{44.8} & \baseline{66.4}\\
& \baseline{Ours w/ NMS} & \baseline{\cmark} &
\baseline{\textbf{83.3}} & \baseline{\textbf{78.4}} & \baseline{\textbf{71.3}} & \baseline{\textbf{60.7}} & \baseline{\textbf{45.6}} & \baseline{\textbf{67.9}}\\

\noalign{\smallskip}
\Xhline{2\arrayrulewidth}

\end{tabular}
}

\vspace{-0.2cm}
\caption{Performance comparison with state-of-the-art methods on THUMOS14. $^*$ and $^\dagger$ denote TSN \cite{TSN} and Swin-B \cite{VideoSwin} backbones, respectively. $^\ddagger$ represents R(2+1)D \cite{R2+1D}.
Others employ I3D \cite{I3D} backbone. The symbol $\mathbf{\Delta}$ indicates partial adoption of NMS.}
\vspace{-0.3cm}
\label{table:thumos}

\end{table}

\subsection{Main Results}

\noindent \textbf{THUMOS14}
Table~\ref{table:thumos} provides a comparison with the state-of-the-art methods on THUMOS14.
Our \modelname{} shows a significant margin of improvement over other query-based detectors, even without NMS.
While TadTR achieves an average mAP of 56.7\% with partial NMS utilization through their proposed cross window fusion (CWF) denoted by $\mathbf{\Delta}$, our method without NMS achieves a superior performance average mAP of 66.4\%.
This significant improvement indicates the length-invariant capability of our \modelname{} model because THUMOS14 contains extremely diverse lengths of videos.
Furthermore, even compared to anchor-free detectors, our method demonstrates competitive performance.

\begin{table}[!t]
\centering
\resizebox{0.9\linewidth}{!}{
\begin{tabular}{ccccccc}
\Xhline{2\arrayrulewidth}
\noalign{\smallskip}
\multirow{2.7}{*}{\textbf{Type}} & \multirow{2.7}{*}{\textbf{Method}} & \multirow{2.7}{*}{\textbf{Feature}} & \multicolumn{4}{c}{\textbf{mAP}}\\
\noalign{\smallskip}
\cline{4-7}
\noalign{\smallskip}
& & & 0.5 & 0.75 & 0.95 & Avg.\\
\noalign{\smallskip}
\Xhline{2\arrayrulewidth}
\noalign{\smallskip}

\multirow{6}{*}{\shortstack{Anchor\\-based}}
& BSN \cite{BSN} & TSN \cite{TSN} &
46.5 & 30.0 & 8.0 & 30.0\\
& BMN \cite{BMN} & TSN \cite{TSN} &
50.1 & 34.8 & 8.3 & 33.9\\ 
& BC-GNN \cite{BC-GNN} & TSN \cite{TSN} &
50.6 & 34.8 & 9.4 & 34.3\\
& G-TAD \cite{G-TAD} & TSN \cite{TSN} &
50.4 & 34.6 & 9.0 & 34.1\\
& VSGN \cite{VSGN} & TSN \cite{TSN} &
52.4 & 36.0 & 8.4 & 35.1\\
& TCANet \cite{TCANet} & TSN \cite{TSN} &
52.3 & 36.7 & 6.9 & 35.5\\

\noalign{\smallskip}
\hline
\noalign{\smallskip}

\multirow{5}{*}{\shortstack{Anchor\\-free}}
& AFSD \cite{AFSD} & I3D \cite{I3D} &
52.4 & 35.3 & 6.5 & 34.4\\
& TALLFormer \cite{TALLFormer} & Swin-B \cite{VideoSwin} &
54.1 & 36.2 & 7.9 & 35.6\\
& ActionFormer \cite{ActionFormer} & R(2+1)D \cite{R2+1D} &
\textbf{54.7} & 37.8 & 8.4 & 36.6\\
& TriDet \cite{TriDet} & R(2+1)D \cite{R2+1D} &
\textbf{54.7} & 38.0 & 8.4 & 36.8\\
& MENet \cite{MENet} & R(2+1)D \cite{R2+1D} & \textbf{54.7} & \textbf{38.4} & \textbf{10.5} & \textbf{37.7}\\

\noalign{\smallskip}
\hline
\noalign{\smallskip}

\multirow{5}{*}{\shortstack{Query\\-based}}
& RTD-Net \cite{RTD-Net} & TSN \cite{TSN} &
47.2 & 30.7 & 8.6 & 30.8\\
& ReAct \cite{ReAct} & TSN \cite{TSN}&
49.6 & 33.0 & 8.6 & 32.6\\
& TadTR \cite{TadTR} & R(2+1)D \cite{R2+1D} &
53.6 & 37.5 & \textbf{10.6} & 36.8\\
& \baseline{\modelname{} (Ours)} & \baseline{R(2+1)D \cite{R2+1D}} &
\baseline{54.0} & \baseline{\textbf{38.2}} & \baseline{\textbf{10.6}} & \baseline{37.0}\\
& \baseline{Ours w/ NMS} & \baseline{R(2+1)D \cite{R2+1D}} &
\baseline{\textbf{54.2}} & \baseline{38.1} & \baseline{\textbf{10.6}} & \baseline{\textbf{37.1}}\\

\noalign{\smallskip}
\Xhline{2\arrayrulewidth}
\noalign{\smallskip}

\end{tabular}
}
\vspace{-0.2cm}
\caption{Comparison with state-of-the-art methods on ActivityNet~v1.3.}
\vspace{-0.6cm}
\label{table:activitynet}
\end{table}
\noindent \textbf{ActivityNet~v1.3}
Following the conventional approach \cite{ActionFormer,TriDet,TadTR}, the external classification score is used to evaluate ActivityNet~v1.3.
The pre-extracted classification scores are combined with predictions from binary detectors to obtain class labels.
Table~\ref{table:activitynet} presents a performance comparison of our \modelname{} with state-of-the-art approaches.
While our \modelname{} method exhibits a slight improvement in query-based detectors on ActivityNet~v1.3 compared to the significant gains shown in Table~\ref{table:thumos}, this is reflective of the intrinsic characteristics of the dataset.
ActivityNet~v1.3 does not contain the diverse length of the video relative to THUMOS14.
Despite this different condition, our approach still demonstrates an improvement on ActivityNet~v1.3, showing performance improvement even though it does not align with the primary issues our \modelname{} aims to resolve.

\begin{table}[!t]
\centering

\resizebox{0.9\linewidth}{!}{
\begin{tabular}{ccccccccc}
\Xhline{2\arrayrulewidth}
\noalign{\smallskip}
\multirow{2.7}{*}{\textbf{Task}} & \multirow{2.7}{*}{\textbf{Method}} & \multirow{2.7}{*}{\textbf{NMS}} & \multicolumn{6}{c}{\textbf{mAP}}\\
\noalign{\smallskip}
\cline{4-9}
\noalign{\smallskip}
& & & 0.1 & 0.2 & 0.3 & 0.4 & 0.5 & Avg.\\
\noalign{\smallskip}
\Xhline{2\arrayrulewidth}
\noalign{\smallskip}

\multirow{8}{*}{Verb}
& BMN \cite{BMN} & \cmark & 
10.8 & 8.8 & 8.4 & 7.1 & 5.6 & 8.4\\
& G-TAD \cite{G-TAD} & \cmark & 
12.1 & 11.0 & 9.4 & 8.1 & 6.5 & 9.4\\
& ActionFormer \cite{ActionFormer} & \cmark & 
26.6 & 25.4 & 24.2 & 22.3 & 19.1 & 23.5\\
& ASL \cite{ASL} & \cmark & 27.9 & - & 25.5 & - & 19.8 & 24.6\\
& TriDet \cite{TriDet} & \cmark & 
\textbf{28.6} & \textbf{27.4} & \textbf{26.1} & \textbf{24.2} & \textbf{20.8} & \textbf{25.4}\\
 & \baseline{\modelname{} (Ours)} & \baseline{\xmark} & 
\baseline{27.0} & \baseline{25.9} & \baseline{24.6} & \baseline{22.9} & \baseline{\textbf{20.0}} & \baseline{24.1}\\
& \baseline{Ours w/ NMS} & \baseline{\cmark} & 
\baseline{\textbf{27.9}} & \baseline{\textbf{26.8}} & \baseline{\textbf{25.4}} & \baseline{\textbf{23.4}} & \baseline{\textbf{20.0}} & \baseline{\textbf{24.7}}\\

\noalign{\smallskip}
\hline
\noalign{\smallskip}

\multirow{8}{*}{Noun}
& BMN \cite{BMN} & \cmark & 
10.3 & 8.3 & 6.2 & 4.5 & 3.4 & 6.5\\
& G-TAD \cite{G-TAD} & \cmark & 
11.0 & 10.0 & 8.6 & 7.0 & 5.4 & 8.4\\
& ActionFormer \cite{ActionFormer} & \cmark & 
25.2 & 24.1 & 22.7 & 20.5 & 17.0 & 21.9\\
& ASL \cite{ASL} & \cmark & 26.0 & - & 23.4 & - & 17.7 & 22.6\\
& TriDet \cite{TriDet} & \cmark & 
\textbf{27.4} & \textbf{26.3} & \textbf{24.6} & \textbf{22.2} & \textbf{18.3} & \textbf{23.8}\\
& \baseline{\modelname{} (Ours)} & \baseline{\xmark} & 
\baseline{26.0} & \baseline{24.8} & \baseline{23.2} & \baseline{20.8} & \baseline{\textbf{18.3}} & \baseline{22.6}\\
& \baseline{Ours w/ NMS} & \baseline{\cmark} & 
\baseline{\textbf{26.3}} & \baseline{\textbf{25.2}} & \baseline{\textbf{23.2}} & \baseline{\textbf{21.0}} & \baseline{18.2} & \baseline{\textbf{22.8}}\\

\noalign{\smallskip}
\Xhline{2\arrayrulewidth}
\noalign{\smallskip}

\end{tabular}
}
\vspace{-8pt}
\caption{Comparison with state-of-the-art methods on EpicKitchens. All methods employ SlowFast \cite{SlowFast} as a backbone.}
\vspace{-10pt}
\label{table:epickitchens}
\end{table}
\noindent \textbf{EpicKitchens}
Table~\ref{table:epickitchens} shows the comparison with state-of-the-art methods on EpicKitchens.
Our model achieves competitive performance without relying on NMS, indicating \modelname{} robustness in diverse and complex action detection scenarios.
EpicKitchen contains an extremely diverse length of action instances, like THUMOS14.
The results indicate the robustness of \modelname{} in handling a wide range of action lengths and complexities.
The comparable performance of \modelname{} is meaningful in query-based approaches, a relatively less explored field than the more established anchor-free methods.

\subsection{Further Analysis}
\vspace{-5pt}

\begin{table}[!t]
\centering

\resizebox{0.75\linewidth}{!}{
\begin{tabular}{cccc}
\Xhline{2.5\arrayrulewidth}
\noalign{\smallskip}
\textbf{Type} & \textbf{Method} & \textbf{NMS} & \textbf{mAP@AVG}\\

\noalign{\smallskip}
\Xhline{2.5\arrayrulewidth}
\noalign{\smallskip}

\multirow{4.5}{*}{Anchor-free} & \multirow{2}{*}{ActionFormer \cite{ActionFormer}} & \xmark & 43.2 (-23.6) \\
& & \cmark & 66.8\\

\noalign{\smallskip}
\cline{2-4}
\noalign{\smallskip}

& \multirow{2}{*}{TriDet \cite{TriDet}} & \xmark & 44.9 (-24.4)\\
& & \cmark & 69.3 \\

\noalign{\smallskip}
\hline
\noalign{\smallskip}

\multirow{7.75}{*}{Query-based}

& \multirow{2}{*}{ReAct \cite{ReAct}} & \xmark & 19.8 (-35.7)\\
& & \cmark & 55.0\\

\noalign{\smallskip}
\cline{2-4}
\noalign{\smallskip}

& \multirow{2}{*}{TadTR \cite{TadTR}} & \xmark & 53.1 (-3.6)\\
& & $\mathbf{\Delta}$ & 56.7\\

\noalign{\smallskip}
\cline{2-4}
\noalign{\smallskip}

&\multirow{2}{*}{\modelname{} (Ours)} & \xmark & 66.4 (-1.5)\\
& & \cmark & 67.9 \\

\noalign{\smallskip}
\Xhline{2.5\arrayrulewidth}

\end{tabular}
}
\vspace{-0.2cm}
\caption{Effect of NMS on the mAP across various anchor-free and query-based methods on THUMOS14. The value in parentheses represents the decrease in mAP when NMS is not applied. The symbol $\mathbf{\Delta}$ indicates partial adoption of NMS.}
\vspace{-0.7cm}
\label{table:nms}
\end{table}

\noindent \textbf{Impact of NMS}
In Table~\ref{table:nms}, we evaluate how NMS influences the performance of various TAD methods on THUMOS14.
NMS is particularly crucial for anchor-free detectors, which employ a one-to-many assignment strategy that leads to duplicated predictions for the same instance.
Furthermore, even though ReAct \cite{ReAct} is a query-based approach, removing NMS at the ReAct significantly affects the performance.
This is why ReAct adopts partial self-attention in the decoder called relational queries, which does not address whole queries.
As shown in Table~\ref{table:attention}, our approach also drops the performance without NMS when removing the decoder's self-attention layer.
This result indicates that addressing whole queries by the decoder's self-attention is crucial to preserving the set-prediction mechanism and full end-to-end modeling.
Furthermore, our proposed method exhibits a minimal performance decrement of only -1.5 when NMS is excluded.
This indicates that our approach effectively achieves full end-to-end modeling.

\begin{table}[!t]
    \centering
    \resizebox{0.72\linewidth}{!}{
        \begin{tabular}{c | cc | cc | c}
            \Xhline{2\arrayrulewidth}
            \noalign{\smallskip}
            Baseline & Multi-Scale & Two-stage [35] & TE & AQS  & mAP@AVG\\
            \noalign{\smallskip}
            \Xhline{2\arrayrulewidth}
            \noalign{\smallskip}
            \multirow{4}{*}{\shortstack{TadTR-34.1s\\ w/ NMS}}
            & & & & & 56.7\\
            & \cmark & & & & 56.5\\
            &  &\cmark & & & 57.3\\
            & \cmark & \cmark& & & 57.0\\
            \noalign{\smallskip}
            \hline
            \noalign{\smallskip}
            \multirow{7}{*}{\shortstack{TadTR-$\infty$\\ w/o NMS}}
            & & & & & 40.2\\
            & \cmark & & & & 42.6\\
            & \cmark & \cmark & & & 43.3\\
            & & & \cmark & & 59.5\\
            & & \cmark &  & \cmark & 46.1\\
            & \cmark & \cmark & \cmark &  & 63.6\\
            & \cmark & \cmark & \cmark & \cmark & 66.4 \\
            \noalign{\smallskip}
            \Xhline{2\arrayrulewidth}
        \end{tabular}
    }
    \vspace{-7pt}
    \caption{Analysis of contributions of each component on THUMOS14. The all-empty check is the denoted baseline.}
    \vspace{-10pt}
    \label{table:contribution}
\end{table}

\noindent \textbf{Component Contribution Analysis}
Table~\ref{table:contribution} shows the incremental impact of each key component in our \modelname{} on THUMOS14.
The performance is measured without NMS.
We conduct experiments based on full end-to-end TadTR, referred to as TadTR-$\infty$ in \cref{subsec:instability_sensitivity}.
Starting with the TadTR-$\infty$ baseline, incorporating multi-scale features and a two-stage approach shows slight improvements.
Nonetheless, these adoptions of improved methods for query-based detectors do not reach the performance level of the original TadTR.
However, including our time-aligned expression (TE) significantly improves performance, demonstrating the value of our time-aligned representations.
Finally, incorporating the adaptive query selection (AQS) mechanism shows the highest performance.

\begin{table}[!t]
    \centering
    \resizebox{0.68\linewidth}{!}{
        \begin{tabular}{c|c|cc|cc}
            \Xhline{2\arrayrulewidth}
            \noalign{\smallskip}
            &\textbf{Encoder} & \multicolumn{2}{c}{\textbf{Decoder}} & \multicolumn{2}{|c}{\textbf{mAP@AVG}}\\
            \noalign{\smallskip}
            \hline
            \noalign{\smallskip}
            &Self-attn. & Self-attn. & Cross-attn. & w/o NMS & w/ NMS\\
            \noalign{\smallskip}
            \Xhline{2\arrayrulewidth}
            \noalign{\smallskip}
            \#1& & \cmark & \cmark & 61.2 & 63.8 \\
            \#2&\cmark &  & \cmark & 53.0 & 63.4 \\
            \#3&\cmark & \cmark &  & 0.1 & 0.2 \\
            \noalign{\smallskip}
            \hline
            \noalign{\smallskip}
            \#4&\cmark & \cmark & \cmark & 66.1 & 67.7 \\
            \noalign{\smallskip}
            \Xhline{2\arrayrulewidth}
        \end{tabular}
    }
    \vspace{-0.2cm}
    \caption{Analysis of the specific roles of self-attention and cross-attention layers in the encoder and decoder, and their impact with and without utilizing NMS.}
    \vspace{-12pt}
    \label{table:attention}
\end{table}
\noindent \textbf{Role of Each Attention}
To clarify the role of encoder and decoder architecture, we conduct experiments for removing the attention layers of the encoder and decoder.
Table~\ref{table:attention} shows the individual contributions of the encoder and decoder layers' attention.
As shown in Row \#1, removing self-attention in the encoder slightly decreases the detection performance, indicating the encoder's self-attention affects representational ability.
Row \#2 reveals that the decoder's self-attention is the core role of the set-prediction mechanism by showing the performance degradation without using NMS.
Finally, Row \#3 shows that removing the decoder's cross-attention does not work because only location information is provided to the decoder, which does not capture the content information without cross-attention.

\begin{figure}[!t]
\centering
    \includegraphics[width=0.65\linewidth]{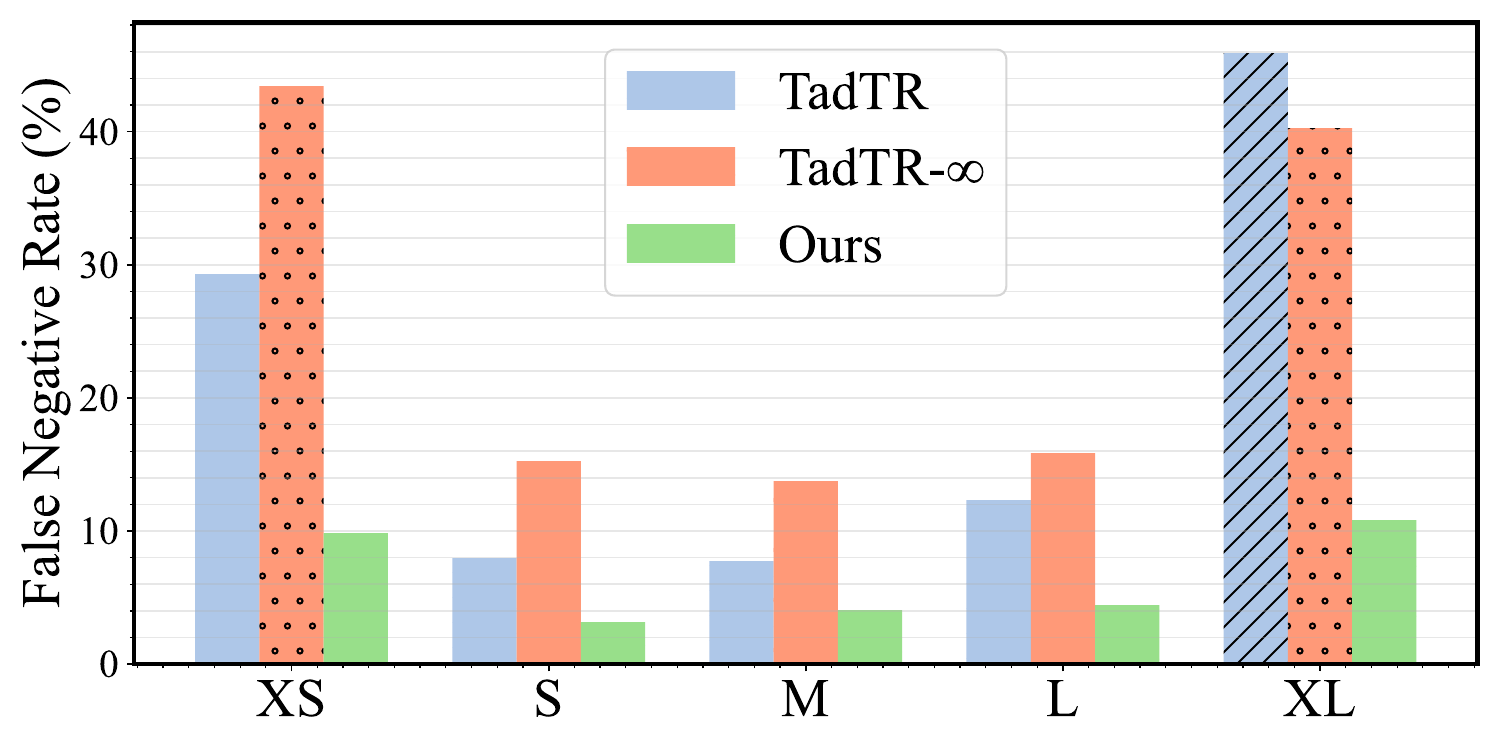}\\
    \vspace{-0.4cm}
    \caption{Comparison of false negative rates by instance lengths (XS, S, M, L, XL) on THUMOS14.}
    \vspace{-0.75cm}
    \label{fig:false_negative}
\end{figure}

\noindent \textbf{False Negative Analysis}
To further compare with the baseline method, we evaluate the false negative rate on THUMOS14.
Fig.~\ref{fig:false_negative} shows the false negative rates across varying action instance sizes: extra small (XS), small (S), medium (M), large (L), and extra large (XL) based on DETAD \cite{DETAD}.
These results show that even though TadTR-$\infty$ shows the worse overall performance in mAP, TadTR-$\infty$ more captures the XL cases, indicating the shorter feature coverage cannot capture the long duration of instances.
These results show that our method significantly reduces false negatives, particularly in XS and XL cases.

\begin{figure}[!t]
\centering
    \includegraphics[width=\linewidth]{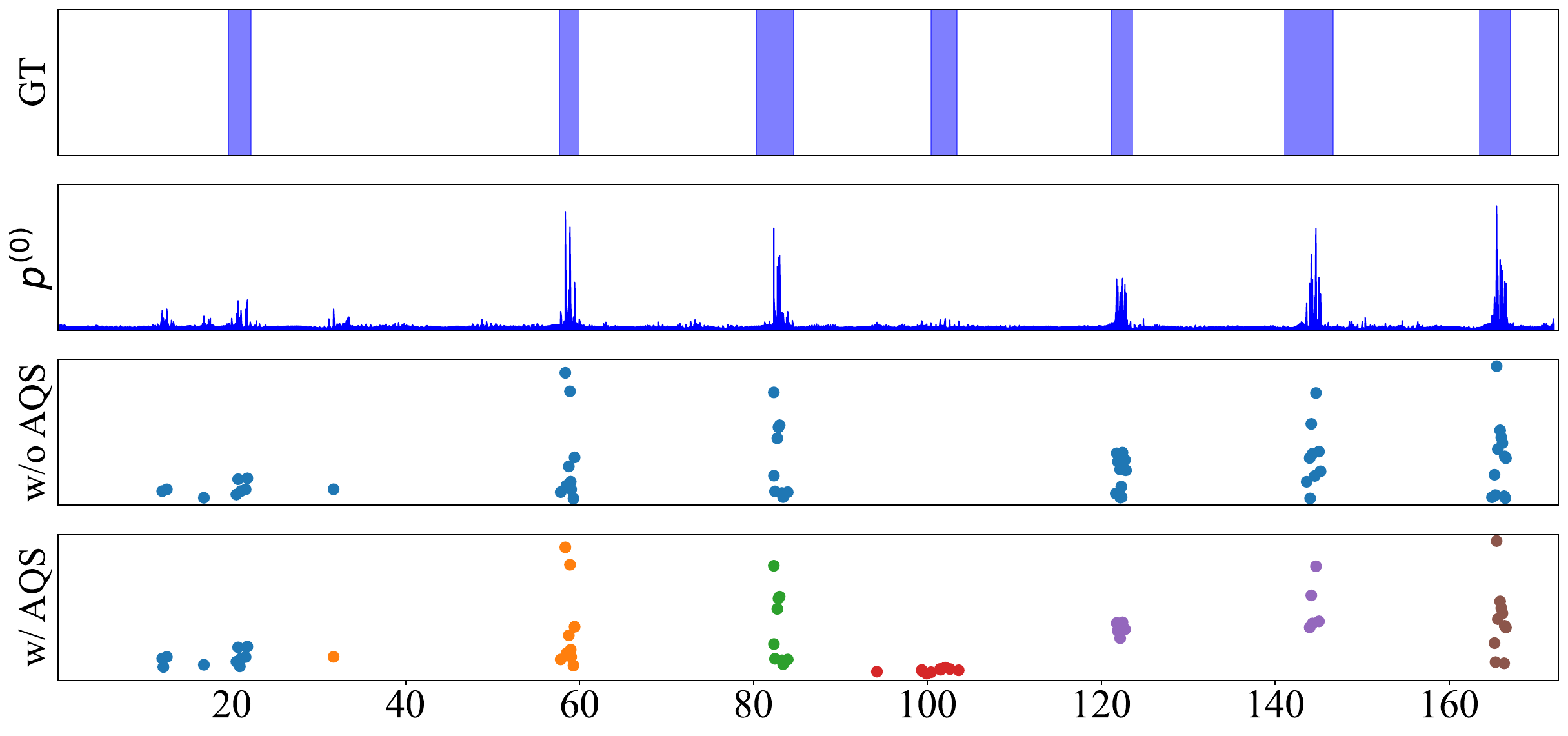}\\
    \vspace{-0.4cm}
    \caption{Comparison between query selection method. Ground truth (GT) is illustrated in the first row. The second row shows encoder prediction scores $p^{(0)}$. The third and fourth row visualizes the generated queries at the top 50 points and 10 points for each sector by AQS, respectively. Different colors in the fourth row represent proposals selected by different sectors.}
    \label{fig:AQP}
    \vspace{-0.7cm}
\end{figure}

\noindent \textbf{Effetiveness of Adaptive Query Selection}
Fig.~\ref{fig:AQP} compares the query selection method by denoting the center point $c^{\text{ref}}$ of selected queries.
Selected points both w/o AQS and w/ AQS are generated by the same encoder output scores.
As illustrated in Fig.~\ref{fig:AQP}, the fixed top-$k$ proposal method misses the fourth ground location as a foreground candidate.
When a foreground proposal is too distant from actual action instances, it necessitates extensive refinement from the decoder.
Refining the segments from missed candidates leads to an additional workload for the decoder layers.
In contrast, our AQS method successfully captures the part missed by w/o AQS.
By dividing the video into sectors and selecting queries within these local units, AQS dynamically adjusts the number of queries and more accurately detects foreground candidates.
\vspace{-10pt}

\section{Conclusion}
\label{sec:conclusion}
\vspace{-5pt}
In this paper, we propose a full end-to-end temporal action detection transformer that integrates time-aligned coordinate expression, called \modelname{}, which eliminates reliance on hand-crafted components such as the sliding window and NMS.
By aligning coordinate expression with the actual video timeline, our model not only simplifies the detection process but also significantly enhances the performance of query-based detectors.
Furthermore, our \modelname{} has a length-invariant property by combining the proposed time-aligned coordinate expression and adaptive query selection, showing the potential of query-based detectors.

\vspace{3pt}
\footnotesize \noindent \textbf{Acknowledgement} \  This work was supported by Institute of Information \& communications Technology Planning \& Evaluation (IITP) grant funded by the Korea government (MSIT) (No. 2019-0-00079, Artificial Intelligence Graduate School Program (Korea University), No. 2021-0-02068, Artificial Intelligence Innovation Hub, and No. 2022-0-00984, Development of Artificial Intelligence Technology for Personalized Plug-and-Play Explanation and Verification of Explanation).
{
    \small
    \bibliographystyle{ieeenat_fullname}
    \bibliography{main}
}


\end{document}